\documentclass{article}

\usepackage{neurips_2025}

\usepackage[utf8]{inputenc} 
\usepackage[T1]{fontenc}    
\usepackage{hyperref}       
\usepackage{url}            
\usepackage{booktabs}       
\usepackage{amsfonts}       
\usepackage{lineno}         
\usepackage{nicefrac}       
\usepackage{microtype}      
\usepackage{xcolor}         

\usepackage{amsmath}        
\usepackage{graphicx}       
\usepackage{booktabs}       
\usepackage{tabularx}       
\usepackage{array}          
\usepackage{multirow}       
\usepackage{caption}        %
\usepackage{booktabs}       
\usepackage{array}          

\usepackage{newfloat}
\usepackage{listings}
\usepackage{amsfonts}
\usepackage{makecell}
\usepackage{float}
\usepackage{tcolorbox}
\tcbuselibrary{listings, breakable}
\usepackage{amsmath}
\usepackage{amssymb}
\usepackage{mathtools}
\usepackage{bm}
\usepackage{cleveref}       
\usepackage{wrapfig}
\usepackage{subcaption}

\newcommand{\method}{CEED-VLA}
\newcommand{\short}{CEED-VLA}

\title{CEED-VLA\includegraphics[height=1em]{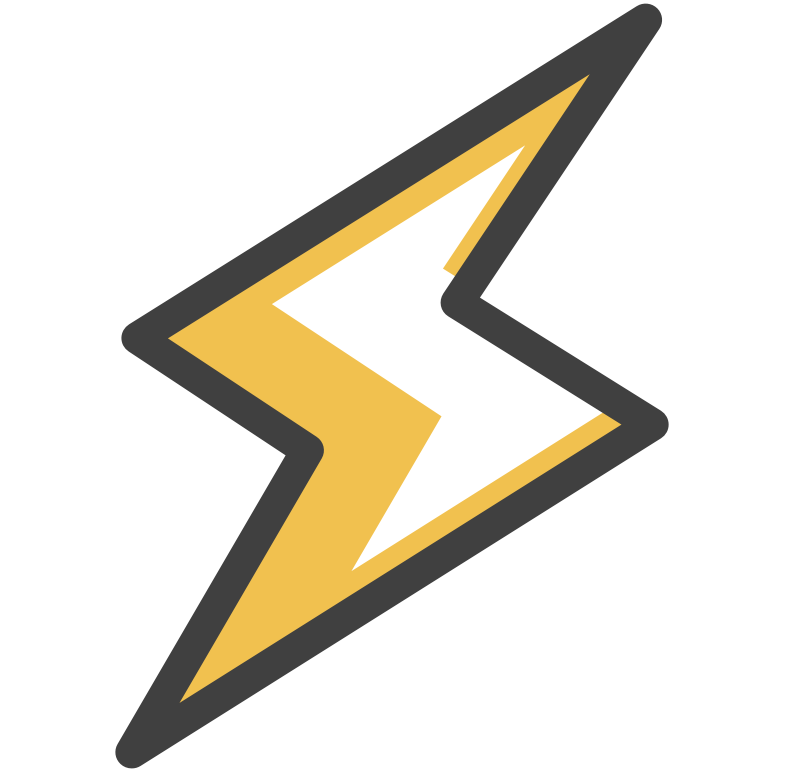}: Consistency Vision-Language-Action Model with Early-Exit Decoding}

\author{Wenxuan Song\textsuperscript{1*}, Jiayi Chen\textsuperscript{1*}, Pengxiang Ding\textsuperscript{2,3\dag}, \\
\textbf{Yuxin Huang\textsuperscript{1}, Han Zhao\textsuperscript{2,3}, Donglin Wang\textsuperscript{2}, Haoang Li\textsuperscript{1\ddag}} \\
\normalsize \textsuperscript{1}HKUST(GZ)
\normalsize \textsuperscript{2}Westlake University
\normalsize \textsuperscript{3}Zhejiang University}

\usepackage{amsmath,amsfonts,bm}

\def\eqref#1{equation~\ref{#1}}

\def\1{\bm{1}}

\def\vx{{\bm{x}}}

\DeclareMathAlphabet{\mathsfit}{\encodingdefault}{\sfdefault}{m}{sl}
\SetMathAlphabet{\mathsfit}{bold}{\encodingdefault}{\sfdefault}{bx}{n}

\newcommand{\E}{\mathbb{E}}

\begin{document}

\maketitle
\begin{abstract}
  In recent years, Vision-Language-Action (VLA) models have become a vital research direction in robotics due to their impressive multimodal understanding and generalization capabilities. Despite the progress, their practical deployment is severely constrained by inference speed bottlenecks, particularly in high-frequency and dexterous manipulation tasks.
  While recent studies have explored Jacobi decoding as a more efficient alternative to traditional autoregressive decoding, its practical benefits are marginal due to the lengthy iterations.
  To address it, we introduce consistency distillation training to predict multiple correct action tokens in each iteration, thereby achieving acceleration.
  Besides, we design mixed-label supervision to mitigate the error accumulation during distillation.
  Although distillation brings acceptable speedup, we identify that certain inefficient iterations remain a critical bottleneck. To tackle this, we propose an early-exit decoding strategy that moderately relaxes convergence conditions, which further improves average inference efficiency.
  Experimental results show that the proposed method achieves more than 4× inference acceleration across different baselines while maintaining high task success rates in both simulated and real-world robot tasks. 
  These experiments validate that our approach provides an efficient and general paradigm for accelerating multimodal decision-making in robotics.
  Our project page is available at \url{https://irpn-eai.github.io/CEED-VLA/}.
\end{abstract}

\renewcommand{\thefootnote}{}
\footnotetext{*~Equal contribution~~\dag~Project Leader~~\ddag~Corresponding author: haoangli@hkust-gz.edu.cn}
\section{Introduction}
\label{sec:intro}
\vspace{-3mm}
Recent advancements in Vision-Language Models (VLMs)~\cite{awadalla2023openflamingo,liu2024visual} have showcased impressive multimodal understanding capabilities, inspiring the development of Vision-Language-Action (VLA) models~\cite{rt1,rt2, octo_2023,niu2024llarva,song2024germ,kim24openvla,driess2023palm,zhen20243d,bjorck2025gr00t,black2024pi_0,song2025accelerating,rationalvla}. These end-to-end architectures, which are trained on large-scale robotic datasets~\cite{o2024open,fang2024rh20t,shafiullah2023bringing,walke2023bridgedata}, integrate visual perception and language understanding to directly generate executable actions.
Although VLA models generalize well across diverse tasks, their practical deployment is severely limited by inference speed bottlenecks, which hinder efficient execution and the handling of high-frequency, dexterous tasks.
Therefore, our goal is to \textbf{significantly improve inference efficiency of VLAs while retaining the manipulation performance}.


To achieve this goal, recent works~\cite{song2025accelerating,openvlaoft} innovatively reframe autoregressive (AR) decoding as a system of nonlinear equations solved through the \textbf{Jacobi fixed-point iteration method}~\cite{ortega2000iterative}.
Specifically, the Jacobi decoding method begins by randomly initializing the $n$ action tokens in a sequence. 
Then, the action sequence along with the prompt is then iteratively processed by the VLA to refine its predictions. 
Through $k$ successive updates, the $n$-token sequence converges to the fixed point, which is the same as the output produced by AR decoding under a greedy strategy. 
Because Jacobi decoding is parallel and allows the model to predict several correct tokens in each iteration, the number of iterations in Jacobi decoding can be fewer than the forward passes in AR decoding, i.e., $k \leq n$.
This indicates that VLAs with Jacobi decoding can be faster than AR ones theoretically.

However, in practice, Jacobi decoding on standard VLAs~\cite{song2025accelerating} delivers only limited acceleration over AR decoding, with recent studies reporting a modest 1.28$\times$ speedup. 
This limitation arises because VLAs are trained in a strictly autoregressive manner, where models are only exposed to ground-truth prefixes during training. 
As a result, when preceding tokens are incorrect, which is a natural condition in Jacobi decoding, the model struggles to generate accurate predictions.
Due to the lack of this capability, the model usually only predicts the first token correctly in each parallel iteration, resulting in slower convergence of the full token sequence, as shown in \Cref{fig:four_de_compare}.

To address this issue, we propose a \textbf{consistency distillation} process in~\Cref{fig:overview}, which enables the student model to predict several correct tokens in one iteration through the training objective of mapping arbitrary points along the Jacobi trajectory of the teacher model directly to the fixed point.
Nonetheless, two key challenges must be addressed: 1. Relying solely on consistency objectives may undermine the model’s inherent autoregressive generation capability. 
2. Error accumulated in the distillation process can result in unreliable supervision signals. 
To mitigate these challenges, we propose a dual-supervision framework. 
First, we introduce an auxiliary AR loss to preserve the student model’s native manipulation skills by aligning next-token distributions with those of the teacher model, thereby regularizing the distillation process. 
Second, we devise a mixed-label supervision mechanism that adaptively combines two sources of supervision. 
When the teacher model’s action accuracy exceeds a learned threshold, its output is used as the supervisory signal.
Otherwise, the target is replaced with the ground-truth value. 
This adaptive strategy ensures robust student learning by selectively leveraging reliable teacher signals while mitigating the impact of performance degradation during distillation.

\begin{figure*}[t!]
    \centering
    \includegraphics[width=0.99\textwidth]{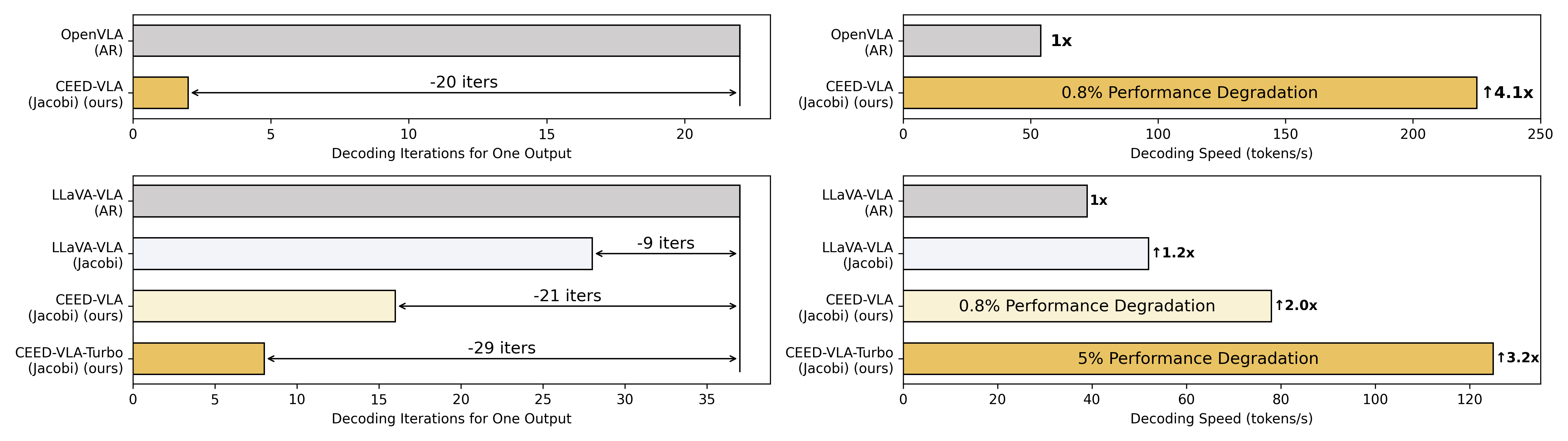}
    \caption{
    \textbf{Acceleration effect of \method~on OpenVLA and LLaVA-VLA.}
    \textbf{Left}: Comparison of the number of iterations required for a complete output. \method~largely reduces the iterations, thus allowing faster decoding. 
    \textbf{Right}: Comparison of the decoding speed. The speedup of directly running Jacobi decoding in a vanilla VLA is marginal.
    Our \method~seperately realizes 3.6$\times$ and 2.0$\times$ speedup with negligible performance degradation on OpenVLA and LLaVA-VLA.
    In scenarios targeting more aggressive acceleration, \method-Turbo delivers even fewer iterations and much more speedup while incurring only a slight degradation in performance.
    }
    \label{fig:small_figure}
    \vspace{-0.5cm}
\end{figure*}

After consistency distillation, the student model demonstrates acceptable acceleration effect, while the ideal acceleration is still bottlenecked by certain \textit{\textbf{inefficient iterations}}.
Due to the strict convergence conditions of Jacobi decoding, these inefficient iterations often require a large number of iterations to reach convergence, significantly reducing the average decoding speed. 
To address this issue, we propose an \textbf{early-exit decoding} strategy, which relaxes the strict convergence conditions of Jacobi decoding and thereby circumvents the impact of inefficient iterations. 
Moreover, our analysis of the structural properties of the task and empirical observation showsthat early-exit decoding has minimal effect on success rates, thus serving as a key accelerator.
Our training approach closely resembles consistency models~\cite{song2023consistency}, as both aim to accelerate inference by directly mapping intermediate equation states to the final solution.
Thus, we term our model as \textbf{C}onsistency \textbf{V}ision-\textbf{L}anguage-\textbf{A}ction model with \textbf{E}arly-\textbf{E}xit \textbf{D}ecoding (\textbf{\method}). 
A variant with a smaller exit point is termed as \method-Turbo, featuring even faster inference.


Experiments in two simulated environments show that
\method~realizes \textbf{2-4.1$\times$}acceleration with comparable success rates across different baselines~(\Cref{tab:baselines}).
Real-world experiments show that \method~realize \textbf{4$\times$} frequency in the real robotic arm deployment with improved success rates on high-frequency dexterous tasks.
We provide a straightforward visualization and detailed analyses, further revealing a key acceleration phenomenon.
We also conducted extensive ablation studies to demonstrate the effectiveness of our key designs and to analyze the impact of training data size.
To summarize, our key technical contributions are as follows:
\vspace{-2mm}
\begin{itemize}
\item We propose \method, a universal acceleration method for significant inference speedup while maintaining manipulation performance.
\item We conduct a consistency distillation process to unlock the model's capabilities of fast inference, and we further propose a mixed-label supervision in the autoregressive loss to preserve the model's manipulation performance.
\item We identify the inefficient iteration as the bottleneck of Jacobi decoding's speedup and propose the early-exit decoding to solve it, resulting in $4.1\times$ speedup, and more than $4.3\times$ frequency.
\end{itemize}

\section{Related Work}
\label{sec:related}
\vspace{-2mm}
\paragraph{Acceleration for Vision-Language-Action Models.}
Various acceleration strategies, including quantization~\cite{lin2024awq} and token pruning~\cite{fastv}, have been effectively applied to LLMs, yet they often fail to meet the stringent real-time requirements of action generation. 
Efforts to enhance efficiency have led to architectural modifications in VLA models, such as DeeR-VLA~\cite{DeeR-VLA}, which dynamically adjusts inference depth, and QAIL~\cite{park2024quantization}, which integrates quantization-aware training. Further innovations, like RoboMamba~\cite{liu2024robomamba} and TinyVLA~\cite{wen2024tinyvla}, replace traditional attention mechanisms or focus on developing lightweight models from the ground up, frequently necessitating model re-training and additional data collection. Meanwhile, VLA-Cache~\cite{vlacache} selectively caches static tokens and recomputes only dynamic or task-relevant ones. 
FAST~\cite{fast} proposes a compression-based tokenization scheme based on the discrete cosine transform.
MoLe-VLA~\cite{zhang2025mole} achieves this by selectively skipping transformer layers based on task-relevant cues.
PD-VLA~\cite{song2025accelerating} framework reformulates autoregressive decoding as a nonlinear system and solves it using a parallel fixed-point iteration method, significantly improving decoding speed while maintaining model performance.
OpenVLA-OFT~\cite{openvlaoft} employs a similar parallel decoding method to speed up.
In contrast, our \method~unlocks the acceleration potential of VLAs by fine-tuning them with a consistency distillation process.
It also optimizes the decoding mechanisms, leading to more significant acceleration.

\paragraph{Consistency Models for Manipulation.}
\vspace{-3mm}
In manipulation tasks, consistency policy~\cite{consistencypolicy} serves as a key acceleration technique for diffusion policies, which employs consistency models~\cite{song2023consistency} in image generation fields to robotics.
It enables single-step inference in a student model while analyzing the impact of design choices like objectives, variance, and chain steps on consistency distillation.
ManiCM~\cite{lu2024manicm} applies consistency constraints to the diffusion process to enable fast inference without compromising action quality.
FlowPolicy~\cite{zhang2025flowpolicy} enables single-step generation by normalizing velocity field consistency, refining flow dynamics for efficient inference.
SDM policy~\cite{jia2024score} integrates score and distribution matching through a dual teacher framework to improve the speed of inference and the quality of action.
While these consistency policies directly map intermediate states of ODEs to their final solution, we train VLAs to map the intermediate points in the Jacobi trajectory to the fixed point.
This paper takes the first step to explore consistency training techniques for efficient VLAs.
\section{Preliminary: Jacobi Decoding}
\label{sec:3}
Given a prompt $\bm{x}$ and a pre-trained LLM $p(\cdot| \bm{x})$, we typically predict tokens using the standard AR decoding method with greedy strategies:
\begin{equation}
\label{eq:ar_decoding}
\begin{aligned}
y_i = \underset{y}{\mathrm{arg\,max}}\ p(y | \mathcal{Y}_{i-1}, \bm{x}) \;\, \text{for}\,\, i = 1,\dots,n,
\end{aligned}
\end{equation}
where $\mathcal{Y}_{i-1}$ denotes $\{y_{1},  \ldots, y_{i-1} \}$, $n$ represents the number of tokens to predict.
Thus, LLM executes $n$ forward passes to obtain $n$ tokens $\mathcal{Y}_n$, which makes it hard to efficiently output a lengthy token sequence. 

Unlike AR decoding, Jacobi decoding~\cite{santilli2023accelerating, cllm} can accelerate the inference by predicting several tokens in one forward.
To directly predict several tokens in each forward, \Cref{eq:ar_decoding} is reformulated as a system of nonlinear equations with respect to $y_i$:
\begin{equation}
\label{eq:jacobi}
\begin{aligned}
\begin{cases}
y_{1}^{(j+1)} &= \underset{y}{\mathrm{arg\,max}}\ p(y |\bm{x}) \\
y_{2}^{(j+1)} &= \underset{y}{\mathrm{arg\,max}}\ p(y | \mathcal{Y}_{1}^{(j)}, \bm{x}) \\
& \vdots \\
y_{n}^{(j+1)} &= \underset{y}{\mathrm{arg\,max}}\ p(y | \mathcal{Y}_{n-1}^{(j)}, \bm{x}),
\end{cases}
\end{aligned}
\end{equation}
where $j$ denotes the $j$-iteration of the Jacobi trajectory.
This enables updates of every token in each forward iteration until convergence.
The nonlinear equation system can be solved in the Jacobi fix-point iteration method~\cite{ortega2000iterative}.
Concretely, Jacobi decoding first initializes a random token sequence $\mathcal{Y}^{(0)}=\{y_{1}^{(0)},  \ldots, y_{n}^{(0)} \}$. 
Then, both the prompt $x$ and the token sequence are fed into the LLM simultaneously.
Then, the variables $y_i$ in $\mathcal{Y}$ are iteratively updated through~\Cref{eq:jacobi} until convergence.
The convergence condition is $\mathcal{Y}^{(k)}=\mathcal{Y}^{(k-1)}$ at the step $k$. 
The $\mathcal{Y}^*:=\mathcal{Y}^{(k)}$ is defined as the \textit{\textbf{fixed point}}. 
Because Jacobi decoding allows to predict multiple correct tokens in the $n$-token sequence in parallel, it requires fewer iterations than AR decoding, thus improving the inference speed.

\section{Method}
\subsection{Teacher Model}
\begin{figure*}[t!]
    \centering
    \includegraphics[width=0.99\textwidth]{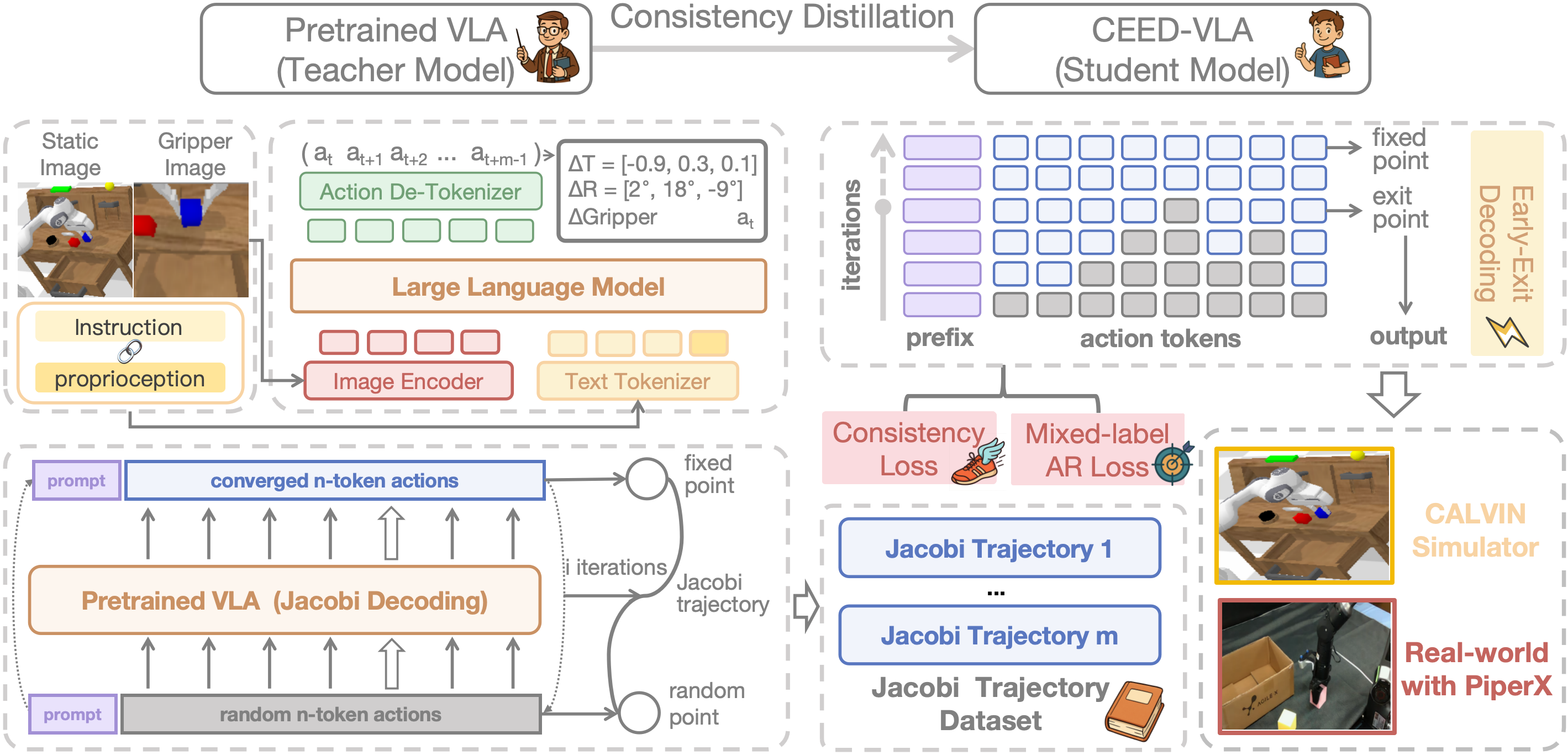}
    \caption{Overview of our proposed \textbf{\method}. Our proposed framework first runs the pretrained VLA (\textit{e.g.}, LLaVA-VLA) with Jacobi decoding to generate the training dataset.
    Then we design an effective consistency distillation process with novel mixed-label supervision to get the student model.
    Finally, we propose early-exit decoding to further unlock inference speed.
    Experiments in simulators and the real world show significant acceleration with comparative success rates.
    }
    \label{fig:overview}
    \vspace{-4mm}
\end{figure*}

\vspace{-2mm}
Vanilla VLAs $P_{\theta}$ learn to predict actions $\hat{a}_{t+1}$ directly from the current observation $s_t$ and language instruction $l$ (\Cref{fig:overview}):
\begin{equation}
    \hat{A}_t \sim P_{\theta}(a_t \mid s_t, l)
\end{equation}

To further improve the planning abilities and stability of actions, the VLAs are combined with action chunking~\cite{act}.
Specifically, at the current time step $t$, given chunk size $m$, the predicted actions will be extended into an action sequences $A_t=[{a_{t+1},a_{t+2},...,a_{t+m}}]$. 
Each $a_t$ consists of 7 dimensions, which is formulated as:
$a_t = [X, \; Y\, \; Z, \; \phi, \; \theta, \; \psi, \; G]$,
where $X,Y,Z$ represent the Cartesian coordinates of the end effector's position, $\phi,\theta,\psi$ denote the rotation angles of the end effector along each axis, and $G$ is the gripper state.
In this paper, we specify 2 vanilla VLA models, the LLaVA-VLA used in~\cite{song2025accelerating, openvlaoft} and OpenVLA~\cite{kim24openvla}, as our teacher models.

Directly replacing the AR decoding in these VLAs with Jacobi decoding really accelerates the inference, while the speedup is \textbf{limited} (1.28$\times$ in~\cite{song2025accelerating}).
The reason is that VLAs, having been trained autoregressively, struggle to produce multiple correct tokens within a single Jacobi iteration: each newly generated token is conditioned on the previously generated ones, so when any preceding token is incorrect, the model is unlikely to generate the correct subsequent token. 




\subsection{Student Model}
\vspace{-2mm}
\label{sec:student model}

 
To address the aforementioned issue and fully unlock the acceleration potential, we finetune VLAs to 
empower them to output multiple correct actions with wrong preceding tokens.
This section first illustrates the data collection for tuning \method~and then details the training procedure.
Finally, we propose an efficient decoding strategy to further accelerate inference.
\vspace{-2mm}

\subsubsection{Jacobi Trajectory Collection} 

\begin{wrapfigure}{r}{0.5\textwidth}  
 \vspace{-1cm}
    \centering
    \includegraphics[width=0.5\textwidth]{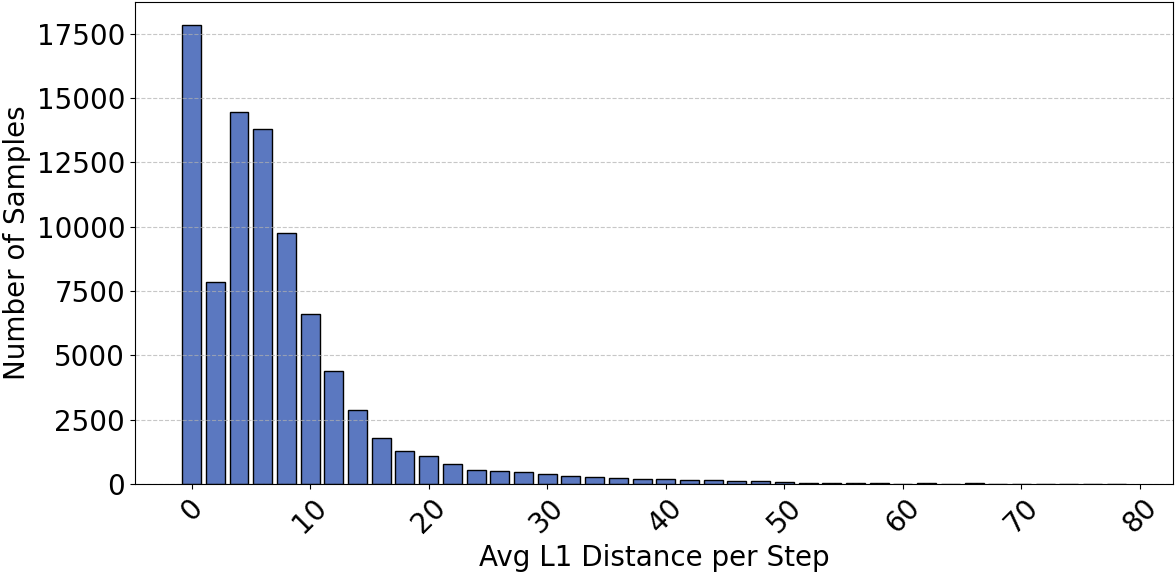}
    \caption{L1 distance between the generated Jacobi trajectory dataset and the ground-truth data.}
    \label{fig:l1_distance}
\vspace{-0.65cm}
\end{wrapfigure}
\label{sec:cvla dataset}
For the target VLA $P$, we let $Q_\theta(\cdot| \vx)$ denotes the \method~with parameters $\theta$ initialized with those of $P$. 
To capture the inherent consistency within Jacobi trajectories, we first collect them by prompting $P$ to 
predict actions with Jacobi decoding on the robot dataset $\mathcal{C}$.
We record the Jacobi trajectories during decoding to build the consistency distillation dataset $\mathcal{D}$. 
The dataset construction process is detailed in \Cref{alg:generate jacobi}.


\subsubsection{Consistency Training}
\vspace{-1mm}

The consistency training procedure optimizes two objectives: \textit{(i)} guiding the model to predict multiple correct tokens simultaneously, and \textit{(ii)} constraining \method~from drifting away from the target VLA distribution to preserve manipulation skills.
\vspace{-3mm}
\paragraph{Consistency Loss.}
The ideal finetuned model consistently maps any point $\mathcal{Y}$ on the Jacobi trajectory $\mathcal{J}$ to the fixed point $\mathcal{Y}^*$. 
We design the consistency loss based on this notion.
Given a prefix $\vx$ consisting of both instructions and visual inputs, and its Jacobi trajectory $\mathcal{J}$, let $\mathcal{Y}$ be a randomly sampled intermediate state and $\mathcal{Y}^*$ the corresponding fixed point.
We train \method~to directly predict $\mathcal{Y}^*$ from $\mathcal{Y}$ by minimizing the following loss:
\begin{equation}
    \label{eq:c_loss}
    \mathcal{L}_{\text{C}} = \E_{ (\vx, \mathcal{J}) \sim \mathcal{D}, \mathcal{Y} \sim \mathcal{J}} \Big[\sum_{i=1}^n \textnormal{KL}\left( Q_{\theta^{-}}(\cdot|\mathcal{Y}_{i}^{*}, \vx) || Q_{\theta}(\cdot|\mathcal{Y}_{i}, \vx) \right) \Big]
\end{equation}
where $\theta^{-} = \text{stopgrad}(\theta)$.
$\textnormal{KL}(\cdot||\cdot)$ denotes the forward KL divergence between two distributions. 

\begin{wrapfigure}{r}{0.45\textwidth}
\vspace{-9mm}
\begin{minipage}{0.445\textwidth}
\input{algorithm/train_consistent_model}
\end{minipage}
\vspace{-0.7cm}
\end{wrapfigure}
\paragraph{Mixed-label AR Supervision.}
To avoid deviating from the distribution of the teacher model, we inherit the AR loss function $\mathcal{L}_{\text{AR}}$ used for training the teacher model:
\begin{equation}
    \label{eq:ar_loss}
    \mathcal{L}_{\text{AR}} = \E_{(\vx, \mathcal{Y}^{*}) \sim \mathcal{D}} \Big[ -\sum_{i=1}^N \log Q_{\theta}(\mathcal{Y}^{*}_{i}|\mathcal{Y}^{*}_{<i}, \vx)\Big].
\end{equation}
To address potential performance degradation stemming from teacher model inaccuracies in action prediction, we implement a mixed-label supervision strategy. 
Specifically, we compute the L1 distance to quantify the distributional divergence between the generated Jacobi trajectory dataset and the corresponding ground-truth action data.
For high-deviation samples, we replace the AR loss labels of those samples with the corresponding ground-truth values.
We define a correctness threshold~$\delta_{\mathrm{max}}$, and any data in the dataset~$\mathcal{D}$ with an L1 distance exceeding this threshold is considered an outlier.
This strategy effectively mitigates error propagation during consistency distillation, maintaining generation quality substantially. 
Consequently, the total loss for training a \method~is:
\begin{equation}
    \label{eq:total_loss}
    \mathcal{L}(\theta) = \mathcal{L}_{\text{C}} + w\mathcal{L}_{\text{AR}}
\end{equation}
where $\omega$ represents a weighting coefficient.
The training procedure is detailed in \Cref{alg:consistent_vla_training}.

\subsubsection{Inference}
\paragraph{Inefficient iterations.}
The student model \method~performs inference via Jacobi decoding, as illustrated in \Cref{sec:3}. 
However, our empirical observations reveal that the acceleration performance of \method~is bottlenecked by a few iterations that are even slower than the minimum speed of AR decoding (see~\Cref{tab:speed_comparison}). 
This phenomenon arises from the \textbf{strict convergence conditions} of the fixed point. 
It requires exact convergence between successive iterations, \textit{i.e.}, $\mathcal{Y}^{(k)} = \mathcal{Y}^{(k-1)}$, which takes numerous iterations (\textit{over 30 iterations}) to reach.
It is further observed that token updates in the final steps of inefficient iterations are marginal, contributing little to the final action decisions.
These inefficient iterations significantly hinder inference speedup and may introduce delays or discontinuities in high-frequency dexterous tasks, ultimately compromising task performance.
\begin{wraptable}{r}{0.45\textwidth}
\vspace{-0.2cm}
\centering
\caption{Comparison of inference speeds among AR decoding, Jacobi decoding, and early-exit decoding on \method.}
\label{tab:speed_comparison}
\begin{tabular}{lccc}
  \toprule
  Decoding & Avg. & Min. & Max. \\ Method &Speed  &Speed  &Speed \\
  \midrule
  AR & 39.64 & 30.87 & 49.10 \\
  Jacobi & 57.57 $\uparrow$ & 19.50 $\downarrow$ & 92.64 $\uparrow$ \\
  Early-exit & 79.24 $\uparrow$& 62.07 $\uparrow$ & 93.53 $\sim$ \\
  \bottomrule
\end{tabular}
\vspace{-0.4cm}
\end{wraptable}
\vspace{-3mm}
\paragraph{Early-exit Decoding.}
To address this limitation, we investigate whether relaxed convergence conditions or fewer iterations are feasible, particularly in terms of their impact on overall performance.
We base our investigation on two aspects:
1. \textbf{Theoretical structural properties}.
Unlike popular VQA tasks (\textit{e.g.} multiple-choice), where each inference has a direct correctness, the success of a robotic manipulation task is decided on a sequence of actions.
Within it, an \textbf{overlooked} yet critical insight is that task success is primarily determined by actions executed at a small number of key states, \textit{e.g.}, releasing the gripper at the target location. 
In contrast, the majority of states along a trajectory do not demand optimal actions, or the set of ``good-enough'' actions is relatively large~\cite{kumar2022should, more}.
Therefore, slight degradation in the quality of certain actions does not adversely affect task outcomes.
2. \textbf{Empirical observation}. We further observe that token updates in the final steps of inefficient iterations are marginal and contribute little to the final action decisions, further highlighting their inefficiency.
This suggests that relaxing convergence and reducing iterations introduce only minor degradation in action quality.
Based on the above analysis and empirical observations, we conclude that relaxing convergence conditions and reducing iterations are feasible without compromising performance.

\begin{wrapfigure}{r}{0.5\textwidth}  
\vspace{-0.5cm}
    \centering
    \includegraphics[width=0.5\textwidth]{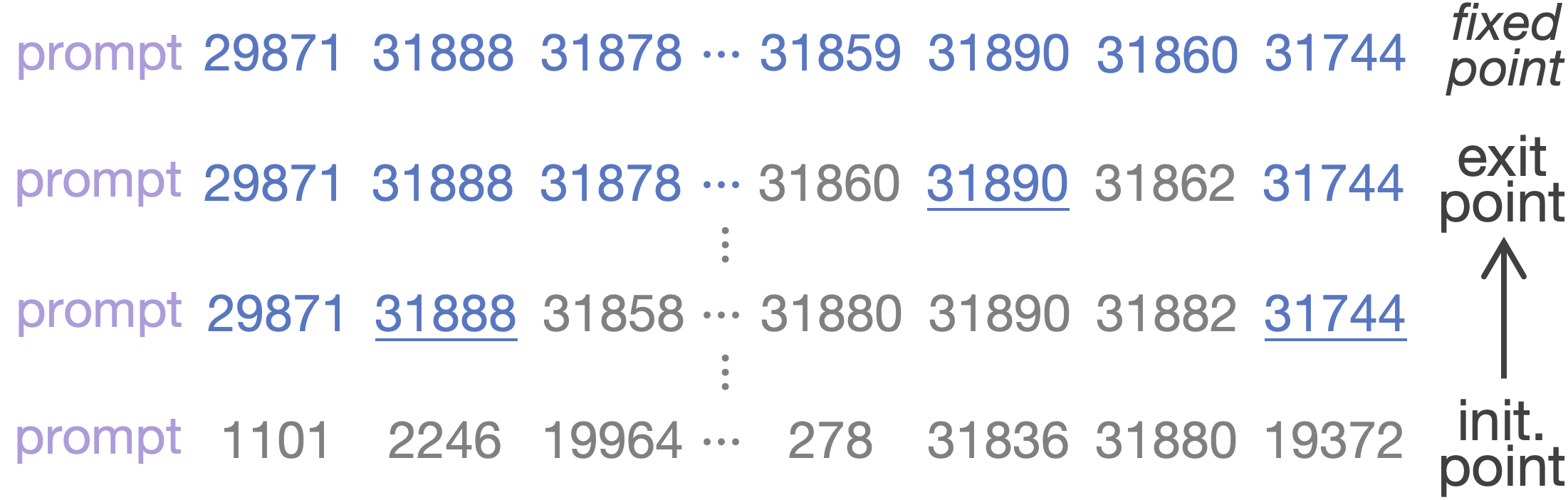}
    \caption{
    \textbf{An instance of Jacobi trajectory with early-exit decoding.}
    Gray numbers indicate incorrect tokens, while blue numbers denote correct ones. Blue numbers with underlines represent fixed tokens.
    The three rows from bottom to top illustrate the Jacobi trajectory, starting from the initialized point and ending at the exit point.
    The topmost row represents the Jacobi fixed point.
}
    \label{fig:vis}
    \vspace{-5mm}
\end{wrapfigure}
To this end, we propose \textbf{early-exit decoding} to relax the condition and significantly reduce the number of iterations. 
Specifically, we introduce the exit point $\sigma$, at which the model prematurely halts the iterative process and directly emits the intermediate output at the $\sigma$-th iteration, bypassing further decoding steps. 
This approach mitigates inefficient iterations, significantly enhancing both the minimum and average decoding speeds while maintaining performance.
We further conduct ablation studies to investigate the optimal value of the exit point, see~\Cref{sec:ablation}.
When the exit point is set to a small value, the model achieves even higher speedup with an acceptable performance drop, termed as \method-Turbo.
\section{Experiments}
\label{sec:exp}
\vspace{-3mm}
In this section, we evaluate the proposed \method~ in terms of its ability to accelerate performance while maintaining manipulation performance. We structure the experiments to answer the following questions: 
\begin{itemize}
    \vspace{-2mm}
    \item  Can \method~serve as a general framework that achieves significant acceleration across different models, tasks, and deployment environments, while maintaining high performance? (see \Cref{sec:5.1})
    \item What \textbf{underlying phenomena} contribute to the observed acceleration effects introduced by \method? (see \Cref{sec:5.1})
    \item How do the \textbf{key design choices} and \textbf{data amounts} in \method~influence the overall model performance and stability? (see \Cref{sec:5.3})
    \item Can \method~effectively increase the execution frequency on \textbf{real-world} robotic platforms, thereby improving the success rate of dexterous manipulation tasks? (see \Cref{sec:5.4})
\end{itemize}
\vspace{-3mm}

\begin{table}[t]
\centering
\small
\caption{Comparison with various manipulation baselines in terms of inference speed, success rates or average length, and execution frequency. All experiments based on LLaVA-VLA are tested on a NVIDIA 4090 GPU and experiments based on OpenVLA are tested on a NVIDIA H100 GPU.} 
\label{tab:baselines}
\scalebox{0.98}{
\begin{tabular}{ccccccc}
\toprule
\multirow{2}{*}{\makecell[c]{Acceleration \\ Techniques}} & \multirow{2}{*}{\makecell[c]{Action \\ Chunking}} & \multirow{2}{*}{\makecell[c]{Decoding \\ Method}} & \multirow{2}{*}{Splits} & Speed & Avg. Len. &Frequency \\
& & &   & (Tokens/s) & SR (\%) &(Hz)\\
\midrule
\multicolumn{7}{c}{Base model: \textbf{LLaVA-VLA} Benchmark: \textbf{CALVIN}} \\
\midrule
N/A & - & AR & ABC$\rightarrow$D & 39.6~\tiny{(\textit{×1})} & 2.01 & 2.23 \tiny{(\textit{×1})}\\ 
N/A & 5  & AR & ABC$\rightarrow$D  & 39.6~\tiny{(\textit{×1})}& \textbf{3.70} & 4.33 \tiny{(\textit{×1.9})}\\
\midrule
FastV& 5 & AR & ABC$\rightarrow$D    & 28.7~\tiny{(\textit{×0.7})}  & 2.54 & 1.87 \tiny{(\textit{×0.8})}\\
SparseVLM& 5 & AR & ABC$\rightarrow$D    & 32.4~\tiny{(\textit{×0.8})}  & 2.83 & 2.01 \tiny{(\textit{×0.9})}\\
PD-VLA & 5 & Jacobi & ABC$\rightarrow$D    & 52.8~\tiny{(\textit{×1.3})} & 3.69 & 5.43 \tiny{(\textit{×2.4})}\\
\midrule
\method~(ours)& 5 & Jacobi & ABC$\rightarrow$D  & \textbf{79.2~\tiny{(\textit{×2.0})}}  & 3.67 & \textbf{7.27~\tiny{(\textit{×3.3})}} \\
\midrule
\multicolumn{7}{c}{Base model: \textbf{OpenVLA} Benchmark: \textbf{LIBERO}} \\
\midrule
N/A& - & AR & Long   & 54.4~\tiny{(\textit{×1})} & 53.2& 5.95~\tiny{(\textit{×1})}\\
N/A& 3 & AR & Long   & 54.4~\tiny{(\textit{×1})} & 60.4& 7.08~\tiny{(\textit{×1})}\\
PD-VLA & 3 & Jacobi & Long    & 85.0~\tiny{(\textit{×1.6})} & \textbf{62.4} & 10.79 \tiny{(\textit{×2.4})}\\
\midrule
\method~(ours)& 3 & Jacobi & Long & \textbf{225.0~\tiny{(\textit{×4.1})}}  & 62.2 & \textbf{25.60~\tiny{(\textit{×4.3})}}\\
\bottomrule
\end{tabular}
}
\vspace{-0.4cm}
\end{table}

\subsection{Simulation Experiments}
\vspace{-5mm}
\label{sec:5.1}

\noindent
\paragraph{Benchmarks and Metrics.}
To evaluate the success rates and inference speedup of our \short, we carefully selected two widely used simulation benchmarks in the robot learning field and VLA tasks for comprehensive experiments.
The \textbf{CALVIN} benchmark~\cite{calvin} includes 34 tasks across four environments (A, B, C, and D). Following the classic CALVIN ABC$\rightarrow$D setup, we run 500 rollouts per model and report success rates and average sequential completions.
\textbf{LIBERO-Long}~\cite{liu2024libero} consists of 10 long-horizon and multi-step manipulation tasks with diverse objects and skills, emphasizing temporal reasoning, goal consistency and delayed rewards.
We evaluate each method across 50 rollouts with varying initial states per task and report both per-task and average success rates. 

\textbf{Baselines.}
In this section, we consider fine-tuning diverse baseline models for a comprehensive validation of our~\short.
\textbf{LLaVA-VLA} is a vanilla VLA model, based on LLaVA~\cite{llava}, detailed in Appendix C.
We also include \textbf{OpenVLA}~\cite{kim24openvla}, the most popular open-source VLA model, for general evaluation.
We further reproduce an OpenVLA with action chunking, which enables more stable actions.
\vspace{-0.3cm}

\paragraph{Acceleration Results in CALVIN.}
We compare our method with na\"ive baselines, advanced acceleration methods for VLMs (FastV~\cite{fastv}, SparseVLM~\cite{zhang2024sparsevlm}), and VLAs in parallel decoding (PD-VLA).
As shown in~\Cref{tab:baselines} , the key findings on CALVIN are as follows: 
1. Only Jacobi decoding without consistency training leads to a modest improvement in VLA decoding speed. 
2. With mixed-label consistency training and early-exit decoding, \method~is able to predict more fixed points, resulting in a 3.2$\times$ speedup, 4.4$\times$ frequencies while the average length only reduces 0.03.

\paragraph{Acceleration Results in LIBERO.}
Experiments on the LIBERO-Long benchmark further validate our conclusions. 
Our proposed \short~ realizes $3.6\times$ speed up and $4.1\times$ frequency, which is more evident than CALVIN.
This is because the tasks in LIBERO are relatively easy and short-horizon, thus requiring fewer iterations in Jacobi decoding.
Meanwhile, the integration with action chunking improves 15.8\% success rates because of stronger planning abilities and smoother actions. 
\begin{table}[t]
\centering
\caption{
Ablation of different techniques used in our~\method~on the base model OpenVLA(action chunk=3).
}
\label{tab:ablation_exp}
\begin{tabular}{c|cccc|cc}
\toprule
\multirow{3}{*}{\makecell[c]{Model}} & \multicolumn{4}{c|}{\makecell[c]{Modifications}} & \multirow{3}{*}{\makecell[c]{Speedup}}  & \multirow{3}{*}{\makecell[c]{Accuracy(\%)}}\\
& \makecell[c]{Jacobi\\Decoding} & \makecell[c]{Consistency\\Training} & \makecell[c]{Mixed-Label\\Supervision} & \makecell[c]{Early-exit\\Decoding} & &\\
\midrule
\method & \checkmark & \checkmark & \checkmark & \checkmark & ×4.1 & 62.2 \\ 
- & \checkmark  & \checkmark & \checkmark  & $\times$ & ×2.5 & 61.2\\
- & \checkmark & \checkmark & $\times$ & $\times$ & ×2.3 & 54.2 \\ 
PD-VLA & \checkmark  & $\times$ & $\times$  & $\times$ & ×1.6 & 62.3\\
OpenVLA & $\times$ & $\times$ & $\times$  & $\times$ & ×1.0 & 60.4 \\ 
\midrule
\end{tabular}
\end{table}
\paragraph{Discuss: Why does OpenVLA achieve more significant acceleration than LLaVA-VLA?}
OpenVLA achieves more than 4× acceleration, in contrast to LLaVA-VLA’s 2× speedup. We hypothesize that this difference stems from the nature of the evaluation tasks: CALVIN features more complex, long-horizon scenarios composed of five subtasks, which demand more careful planning during execution, thereby increasing the number of required iterative steps.

\begin{wraptable}{r}{0.4\textwidth}
\centering
\footnotesize
\caption{Profiling results for the number of fixed tokens and accelerations.}
\label{tab:fixed_token}
\begin{tabular}{lcc}
\hline
Model& \makecell{Num. of\\\textit{fixed token}}   & Speedup  \\
\hline
LLaVA-VLA & 0 & 1$\times$ \\
PD-VLA & 8.75 & 1.33$\times$ \\
\short~ & 13.5 & \textbf{2.00$\times$} \\ 
\hline
\end{tabular}
\vspace{-2mm}
\end{wraptable}
\paragraph{Visualization of Jacobi trajectory with early-exit decoding.}
We randomly visualize a Jacobi trajectory during testing in \Cref{fig:vis}. 
The trajectory starts from a randomly initialized point and iteratively updates until reaching the predefined exit point, where the final output is generated. 
Notably, the value at the exit point does not exactly match the conventional Jacobi fixed point, indicating that the model terminates earlier before full convergence. 
The early-exit mechanism further reduces the number of iterations and contributes to the observed acceleration.

\paragraph{Underlying Acceleration Phenomena.}
In \Cref{fig:vis}, we observe that our \short~is capable of correctly predicting certain action tokens (underlined blue numbers) in one iteration, even if preceding tokens are incorrect. 
We refer to such tokens as \textit{fixed tokens}. 
The presence of fixed tokens significantly enhances the model's ability to predict multiple tokens within each iteration. 
Table~\ref{tab:fixed_token} presents the correlation between the number of fixed tokens and the resulting acceleration gains.
\subsection{Ablation Experiments}
In this section, we first evaluate the significance of several key designs in our method on the base of OpenVLA.
As illustrated in Table~\ref{tab:ablation_exp}, consistency training and early-exit decoding substantially accelerate the model, whereas mixed-label supervision plays a crucial role in maintaining action accuracy and success rate.

Then we investigate the choice of different exit points, data amounts, supervising labels, and losses in turn.
The main model is fine-tuned on LLaVA-VLA with an exit point of 16, 960k training trajectories, mixed-label supervision, and a loss ratio of 1.

\label{sec:ablation}
\vspace{-2mm}
\label{sec:5.3}
\begin{figure}[t]
  \centering
  \begin{subfigure}{0.48\linewidth}
    \includegraphics[width=\linewidth]{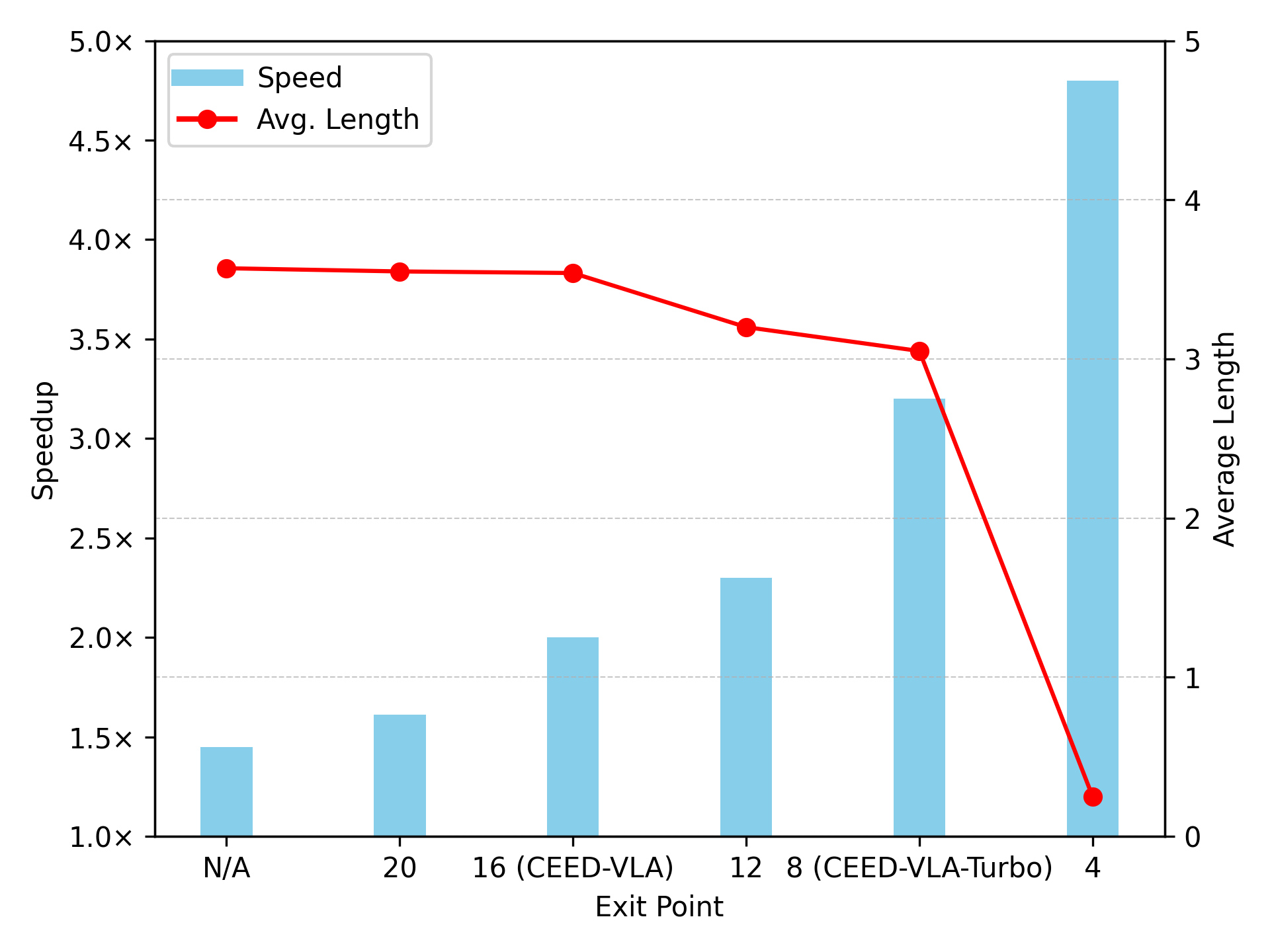}
    \label{fig:img1}
  \end{subfigure}
  \hfill
  \begin{subfigure}{0.48\linewidth}
    \includegraphics[width=\linewidth]{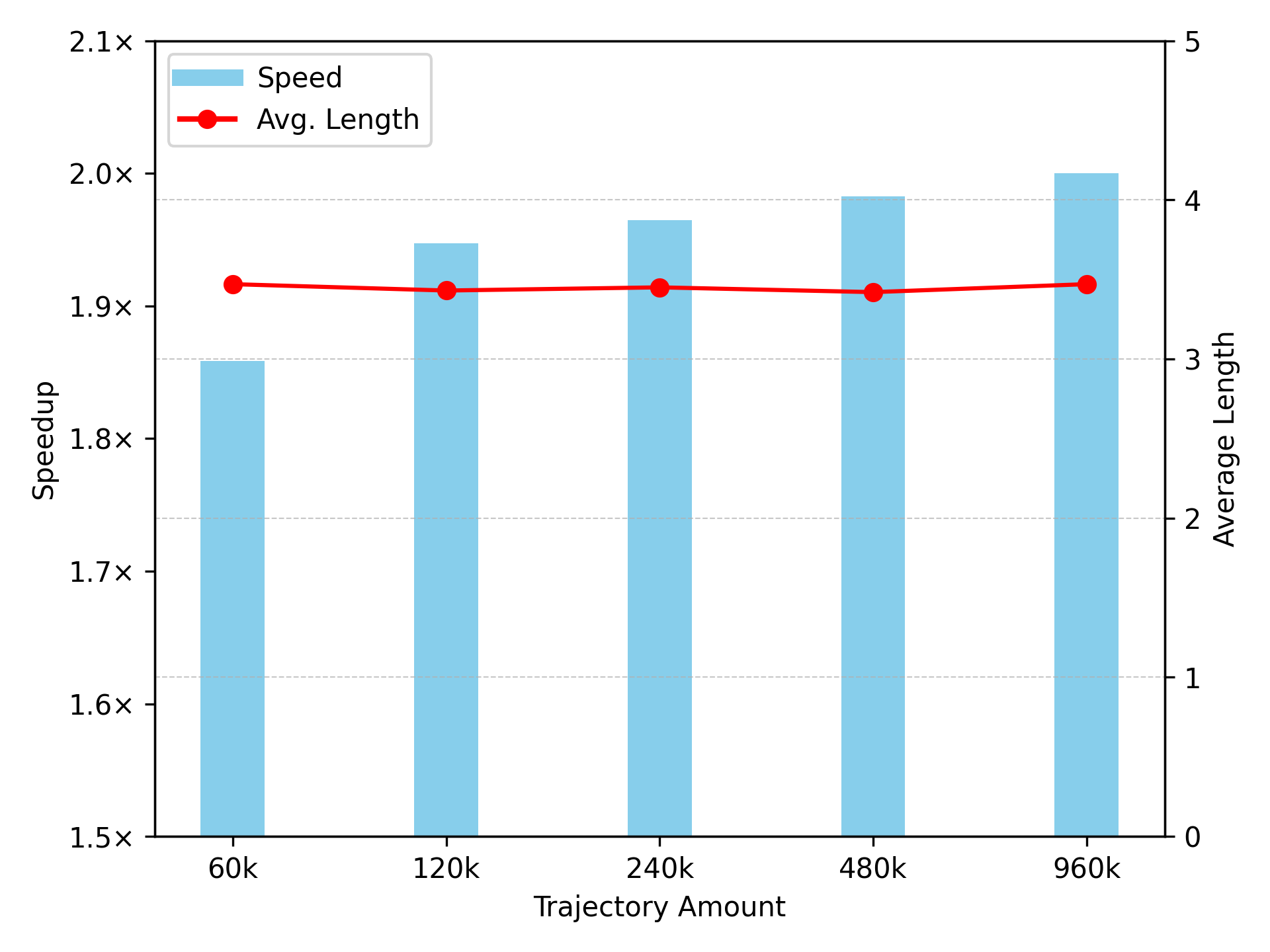}
    \label{fig:img2}
  \end{subfigure}
  \vspace{-0.5cm}
  \caption{
  Speedup and average length of \method~decoding with different values of exit point (\textbf{left}) and trained with different data amounts (\textbf{right}).
  On the left, our \method~employs an exit point of 16, and the extremely accelerated version \method-Turbo exits at 8.
  }
  \label{fig:ablation}
  \vspace{-0.5cm}
\end{figure}

\begin{wraptable}{r}{0.4\textwidth}
\vspace{-3mm}
\centering
\footnotesize
\caption{Comparison of performance between different supervising labels for AR loss.}
\label{tab:mix}
\begin{tabular}{lcc}
\hline
Label & Speedup & Avg. Len. \\
\hline
Mixed (ours) & \textbf{2.00$\times$} & \textbf{3.67} \\
Teacher Model & 1.83$\times$ & 3.48 \\
GT & 1.67$\times$ & 3.20 \\
\hline
\end{tabular}
\end{wraptable}

\noindent
\textbf{Early-exit Decoding.}
Early-exit decoding serves as a key accelerator, balancing inference speed and success rates. In \Cref{fig:ablation}, we conduct inference on different values $\sigma$ of exit points. 
AR decoding requires 37 iterations to complete a single action. 
We observe that introducing early-exit decoding and setting $\sigma$ to 16, \textit{i.e.}, a maximum of 16 iterations, leads to a noticeable speedup with negligible performance drop. 
This suggests that enabling early exit at a few steps within a task does not affect overall success. 
Reducing the $\sigma$ further to 8 results in a slight performance degradation but brings even greater acceleration. 
This version is termed as \method-Turbo.
However, lowering the exit point to 4 yields extreme speedup at the cost of substantial action, making task success difficult.

\noindent
\textbf{Mixed-label Supervision.}
We investigate the impact of label choices for supervising the AR loss.
As shown in \Cref{tab:mix}, our mixed-label supervision best balances between speed-up and performance.
This is because it effectively anchors the student model between the teacher output and the ground truth, preventing drift from either side.
Using only the teacher output as the label causes error accumulation, resulting in shorter average episode lengths.
Conversely, using only the ground truth creates conflicting supervision between the AR and consistency losses, yielding the worst performance.

\begin{wraptable}{r}{0.4\textwidth}
\vspace{-5mm}
\centering
\footnotesize
\caption{Comparison of inference speed and average length between different ratios of losses to evaluate the trade-off.}
\label{tab:quant}
\begin{tabular}{lcc}
\hline
Loss & Speedup & Avg. Len. \\
\hline
$\mathcal{L}_{\text{C}} + \mathcal{L}_{\text{AR}}$(ours) & \textbf{2.00$\times$} & 3.67 \\
$\mathcal{L}_{\text{C}} + 10*\mathcal{L}_{\text{AR}}$ & 1.79$\times$ & \textbf{3.74} \\
\hline
\end{tabular}
\vspace{-4mm}
\label{tab: loss}
\end{wraptable}
\noindent
\textbf{Loss Design.}
The consistency loss guides the model to learn the 
convergence behavior of the Jacobian trajectory toward a fixed point, 
thereby reducing the iterations and accelerating the overall decoding process. 
The autoregressive (AR) loss encourages the model to learn correct actions, preventing it from deviating from the distribution of teacher model and ensuring performance. 
The trade-off between speed and accuracy is influenced by the relative weighting of the two losses. We experimented with weight ratios of 1 and 10, respectively. 
As shown in \Cref{tab: loss}, increasing the emphasis on the AR loss does indeed improve accuracy, albeit at the cost of speedup.


\textbf{Data Size.}
\short~learns to predict multiple tokens in each step by leveraging pre-collected Jacobian trajectories.
As shown in \Cref{fig:ablation}, 60k trajectories are sufficient to achieve significant acceleration, demonstrating strong data efficiency because training for one epoch on this dataset takes only 10 hours.
When the number of trajectories exceeds 120k, the performance gain plateaus and becomes marginal.
Another observation is that increasing the data volume has a limited impact on the average episode length.
\vspace{-2mm}
\begin{figure}
    \centering
    \includegraphics[width=0.935\textwidth]{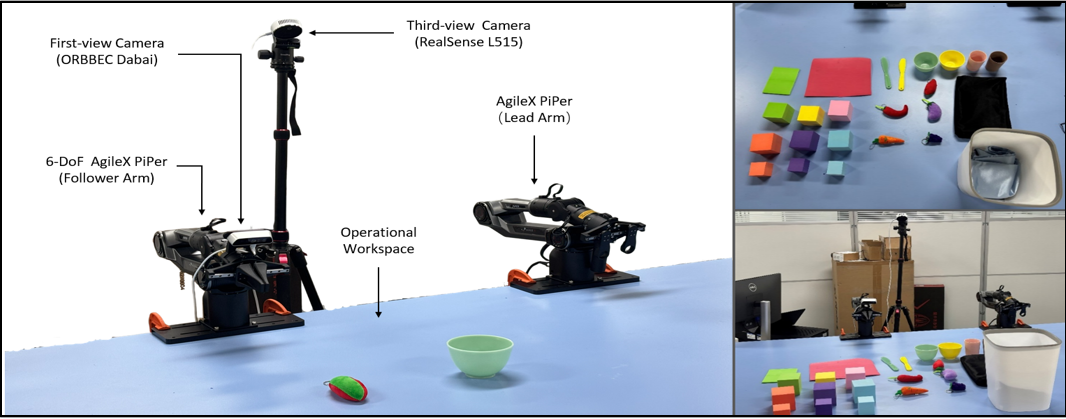}
    \caption{\textbf{The real-world robotic system for experiments}. The system consists of two AgileX PiPer 6-DoF robotic arms, an ORBBEC Dabai depth camera, and a RealSense L515 depth camera.}
    \label{fig:real_setup}
    \vspace{-0.5cm}
\end{figure}
\subsection{Real-world Eexperiments}
\label{sec:5.4}
\noindent
\textbf{Real-world Setup}. 
Our real-world experiments were performed on a dual-arm AgileX PiPer robotic system. 
During the data collection phase, one arm was teleoperated by a human to provide demonstrations (designated as the lead arm), while the other arm (designated as the follower arm) remained passive. During evaluation, only the follower arm was controlled by VLAs to perform the tasks.
To provide visual observations, we employ a RealSense L515 depth camera as Third-view and an ORBBEC Dabai depth camera as First-view input.

\noindent
\textbf{Tasks and Datasets.}
Our dataset covers three levels of manipulation difficulty, categorized into basic Tasks and dexterous Tasks, as described below.
Basic Tasks are short-horizon, atomic interactions such as button pushing and block lifting. These tasks focus on simple object interactions under basic planning and are collected at a frequency of 10 Hz.
Dexterous Tasks demand high-frequency, continuous control, and fine-grained manipulation skills. 
These tasks include towel folding and water pouring, which are collected at 30Hz.
\begin{table}[t]
\centering
\small
\caption{Real-world success rate comparison with LLaVA-VLA. 
We evaluate all the methods with 4 (tasks) × 20 (variants) rollouts. 
Our method achieves better performance among all tasks than baselines.
}
\label{tab:real}
\scalebox{0.98}{
\begin{tabular}{lcccccc}
\toprule
Method & \makecell{Push button\\(basic)} & \makecell{Lift block\\(basic)}  & \makecell{Pour water\\ (dexterous)} & \makecell{Fold towel\\ (dexterous)} & \makecell{Avg. success\\ rate (\%)} & \makecell{Frequency\\(Hz)} \\
\midrule
LLaVA-VLA & 65 & 40 & 15 & 5 & 33.25 & 3.3 \\
PD-VLA    & 85 & 65 & 45 & 35 & 57.50 & 6.0~\tiny{(\textit{1.8$\times$})} \\
\short~(ours)   & \textbf{85} & \textbf{70} & \textbf{80} & \textbf{75} & \textbf{77.50} & \textbf{13.0}~\tiny{(\textit{3.9$\times$})} \\
\bottomrule
\end{tabular}
}
\vspace{-0.5cm}
\end{table}
\vspace{-0.3cm}
\paragraph{Results.}
\Cref{tab:real} presents the real-world results.
We observe that LLaVA-VLA achieves decent success rates on basic tasks, but struggles to learn effective policies with high-frequency robot data. 
Its discontinuous actions often lead to task failures.
PD-VLA improves action continuity and execution frequency by integrating parallel decoding with action chunking, leading to higher success rates on both basic and dexterous tasks.
Our \method~significantly boosts inference speed and control frequency, enabling the model to learn and execute high-frequency actions. 
As a result, it substantially improves success rates on dexterous tasks, exceeding 70\%.

\section{Conclusions}
\label{sec:conclusions}
In this paper, we propose \method, a universal acceleration method for significant inference speedup
while maintaining manipulation performance.
We propose a mixed-label supervision during consistency training and early-exit decoding during inference.
Extensive experiments in several simulation and real-world environments demonstrate that our \method~significantly improves the inference speed and execution frequency, while maintaining comparative manipulation performance and better managing high-frequency dexterous tasks.

\section{Future work}
\label{sec:future}
Recent works investigate more efficient tokenization schemes for robot actions in autoregressive VLAs~\cite{fast, minivla}.
They allow representing $x_2$ action components with $x_1$ tokens where $x_1 < x_2$, thereby improving the efficiency of model training and inference, as well as the capture of high-frequency actions. Combining our method with these works to further achieve real-time inference is a promising direction.
Our method also demonstrates potential to drive progress across embodied chain-of-thought~\cite{ecot} that produce extensive intermediate reasoning tokens and knowledge insulating~\cite{pi_ki} that require simultaneous generation of both linguistic responses and action tokens.

\newpage

\bibliographystyle{IEEEtran}
\bibliography{root}

\newpage
\appendix

\section{Limitations}
\vspace{-1mm}
\label{sec:limitations}
It easily applies to models with discrete actions, but its \textbf{limitation} is less suitable for continuous action spaces where convergence conditions are harder to define.
We will explore more effective and unified convergence conditions in future work.

\section{Generating Jacobi Trajectory Datasets}

\begin{figure}[h]
\begin{minipage}{0.98\textwidth}
    \input{algorithm/generate_jacobi}
\end{minipage}
\end{figure}
\section{Benchmark}
\vspace{-1mm}
We carefully selected two simulation benchmarks that are widely evaluated in robot learning fields and VLA tasks.
The \textbf{CALVIN} benchmark~\cite{calvin} is built on top of the PyBullet~\cite{pybullet} simulator and involves a Franka Panda Robot arm that manipulates the scene. CALVIN consists of 34 tasks and 4 different environments (A, B, C and D). 
We evaluate all methods on the classic \textbf{CALVIN ABCD$\rightarrow$D} setup~\cite{calvin}. 
Each model is tested for 500 rollouts, and we report success rates and the average number of completed sequential tasks.
\textbf{LIBERO-Long}~\cite{liu2024libero} consists of long-horizon tasks, encompassing diverse object interactions and versatile motor skills.
To evaluate a model's ability in temporal reasoning and long-horizon planning, we introduce the \textbf{LIBERO-Long} benchmark. This suite consists of multi-step manipulation tasks that require agents to complete a sequence of subgoals in a coherent and temporally aligned manner. Unlike atomic tasks that focus on single-stage actions, LIBERO-Long challenges the model with increased task complexity, delayed rewards, and the need for consistent goal tracking.
Each method is evaluated across 100 rollouts with varying initial states for each task. We report both per-task and average success rates. 
\section{Baseline}
\begin{figure}[h]
  \centering
    \includegraphics[width=1\linewidth]{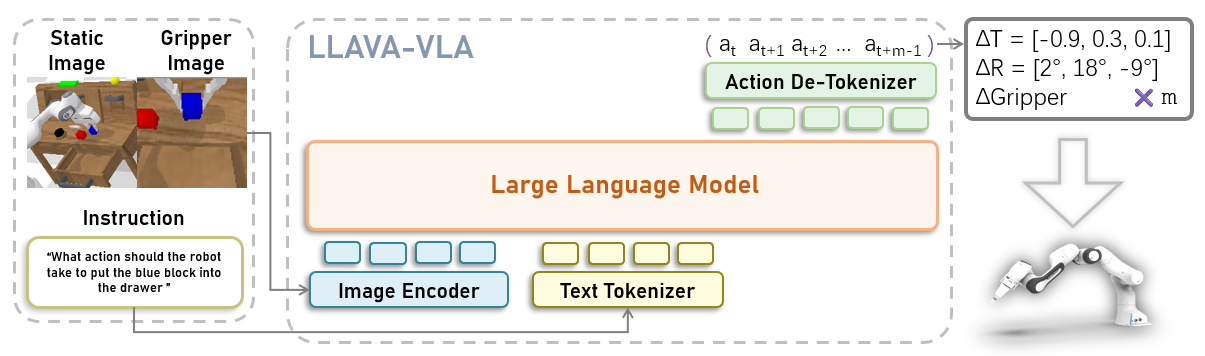}
    \caption{\textbf{LLaVA-VLA model architecture.} Given a sequence of images and language instructions, the input is first tokenized and fed into the language model. The model then generates action tokens, which are subsequently detokenized into executable action values and deployed on the robotic arm.}
  \label{fig:llava-vla}
  \vspace{-2mm}
\end{figure}
In this section, we consider fine-tuning diverse baseline models for a comprehensive validation of our~\short.
\textbf{LLaVA-VLA} is a vanilla Vision-Language-Action (VLA) model with action chunking at the output stage, derived by fine-tuning the widely adopted vision-language model LLaVA~\cite{llava}. An overview of the model architecture is provided in \Cref{fig:llava-vla}.
It incorporates action chunking at the output stage to promote temporally coherent action generation. 
LLaVA-VLA exhibits stable performance across both simulated and real-world environments. It has also been adopted as a baseline in several prior works~\cite{vlas, song2025accelerating}. 
We conduct most experiments based on LLaVA-VLA on the CALVIN ABC-D task to investigate the effectiveness of \short.

\textbf{OpenVLA}~\cite{kim24openvla}, on the other hand, is the most widely used open-source VLA model. The model architecture builds upon an LLaMA 2 language model, augmented with a visual encoder that integrates pretrained features from both DINOv2 and SigLIP, enabling robust multimodal grounding in complex manipulation tasks.
We further validate the general applicability of our approach by using OpenVLA as an additional baseline.
\vspace{-1mm}
\section{Adaptability of Our Proposed Method}
\vspace{-1mm}
Our method introduces a lightweight and practical solution that integrates seamlessly with existing vision-language models. It offers three distinct advantages. First, it is model redesign-free, meaning that no architectural changes to the backbone model are required, thereby preserving the integrity and performance characteristics of the original network. Second, it is modification-free as it does not require modifications or adding auxiliary components to pre-trained VLA models. Third, the method incurs no additional GPU memory overhead, making it highly resource-efficient and suitable for deployment in environments with constrained computational budgets. Together, these features make our approach both flexible and scalable, facilitating broad applicability without sacrificing efficiency or accuracy.
\vspace{-1mm}

\section{Visualization of the Decoding Process}
\begin{figure}[H]
  \centering
    \includegraphics[width=0.995\linewidth]{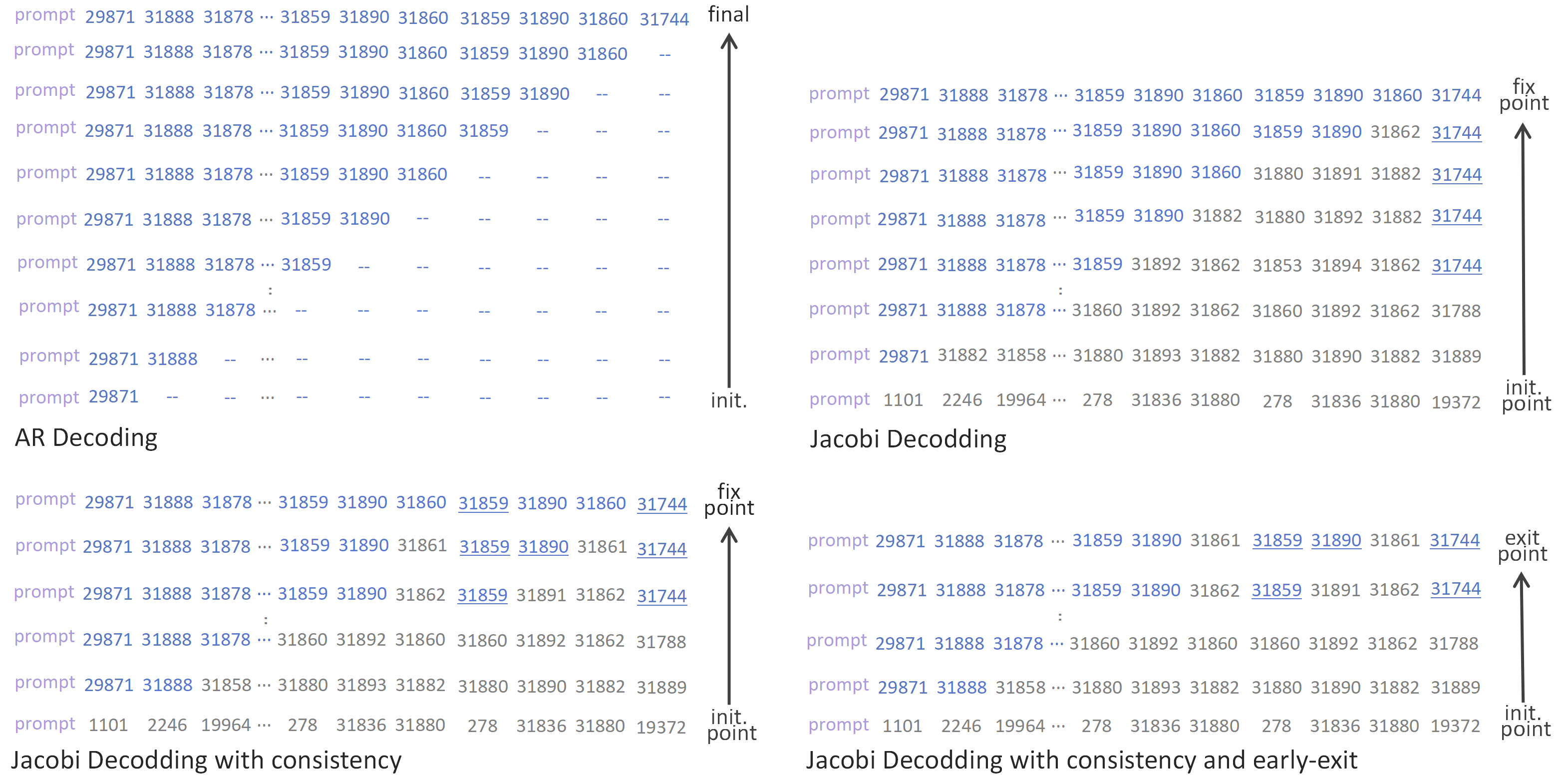}
    \caption{\textbf{Comparation of four decoding methods}}
  \label{fig:four_de_compare}
  \vspace{-2mm}
\end{figure}


\textbf{AR Decoding}. 
As shown in the top-left corner of \Cref{fig:four_de_compare}, AR decoding is the conventional and widely adopted approach in sequence generation tasks. It generates tokens sequentially in a left-to-right manner, where each token depends strictly on all previously generated tokens. This method ensures stable performance and high output accuracy. However, its inherently sequential nature prevents parallelization, leading to slow inference and becoming a major bottleneck in real-time applications.

\textbf{Jacobi Decoding}. 
As shown in the top-right corner of \Cref{fig:four_de_compare}, Jacobi decoding draws inspiration from the classical Jacobi iterative method in numerical computation. It enables simultaneous updates of all tokens by iteratively refining predictions based on the outputs from the previous step.
While this method successfully breaks the sequential dependency of autoregressive decoding and supports parallelism, its convergence speed remains relatively slow. Without specialized training strategies, its acceleration potential is limited in practice.

\textbf{Jacobi Decoding with consistency training }. 
As shown in the Bottom-left corner of \Cref{fig:four_de_compare}, Consistency training is introduced to enhance the model’s ability to converge from arbitrary initial predictions toward a fixed point. By encouraging the prediction of more accurate tokens across iterations, it significantly reduces the number of steps required for convergence, thereby improving decoding efficiency.
Despite this advancement, inference still suffers from inefficiency points, where certain tokens require more iterations to stabilize. These outlier tokens constrain the minimum achievable decoding latency, leaving room for further optimization.

\textbf{Early-exit Decoding with consistency training }. 
As shown in the Bottom-right corner of \Cref{fig:four_de_compare}, building on consistency training, the early-exit decoding with consistency training strategy allows token-wise termination once predictions converge or exhibit negligible change. Alternatively, a global iteration cap can be imposed to force early termination across all tokens.
The empiricalical results shown in \Cref{fig:ablation}, demonstrate that models trained with consistency loss maintain robust performance even under strict iteration limits. Notably, limiting the number of iterations at certain decoding steps does not significantly affect the overall task completion rate.

By striking an elegant balance between speed and performance, early-exit Jacobi decoding represents one of the most practical and efficient solutions for parallel inference.

\section{Ablation experiment}
Consistency training preserves the performance of early-exit decoding.
As shown in Table.~\ref{tab:CT}, we compare the average and average length of models with and without consistency training under different exit points.
Consistency models exhibit more pronounced acceleration effects while maintaining high success rates even with early exit points. This stems from their ability to converge most action tokens with minimal iterations, whereas vanilla models fail to converge adequately under limited iterations, suffering significant performance degradation.

\begin{table}[t]
\centering
\small
\caption{
Evaluation of LLaVA-VLA on the CALVIN ABC→D benchmark, focusing on early-exit behavior. Metrics include speed and accuracy. Experiments are conducted on an NVIDIA RTX 4090 GPU.
}
\label{tab:CT}
\scalebox{0.98}{
\begin{tabular}{cccc}
\toprule
Exit point & Decoding Method & Avg. Speed (Tokens/s) & Avg. Accuracy \\
\midrule
\multicolumn{4}{c}{\textbf{w/o Consistency Training}} \\
\midrule
--             & AR      & 39.6   & 3.50 \\
no (pd-vla)    & Jacobi  & 52.8   & 3.49 \\
16             & Jacobi  & 56.2   & 3.08 \\
12             & Jacobi  & 74.2   & 1.79 \\
8              & Jacobi  & 87.3   & 1.24 \\
\midrule
\multicolumn{4}{c}{\textbf{with Consistency Training}} \\
\midrule
no             & Jacobi  & 57.42  & 3.49 \\
16 (ceed-vla)  & Jacobi  & 79.2   & 3.47 \\
12             & Jacobi  & 89.1   & 3.20 \\
8 (ceed-vla-turbo) & Jacobi  & 126.72 & 3.04 \\
\bottomrule
\end{tabular}
}
\vspace{-0.4cm}
\end{table}


\end{document}